\documentclass[11pt,a4paper]{article}
\usepackage[hyperref]{eacl2021}
\usepackage{times}
\usepackage{latexsym}

\usepackage{graphicx}
\usepackage[ruled, linesnumbered]{algorithm2e}
\usepackage{amssymb,mathtools}
\usepackage{xcolor}
\usepackage{comment}
\usepackage{mathtools}
\usepackage[thinlines]{easytable}
\usepackage{babel,blindtext}

\usepackage{diagbox}
\usepackage{multirow}
\usepackage{longtable}
\usepackage{dblfloatfix}
\usepackage{scrextend}
\usepackage{float}
\usepackage{caption}
\usepackage{url}
\usepackage{multicol}
\usepackage{amsfonts}
\usepackage{subfigure}

\usepackage{array}
\newcolumntype{L}[1]{>{\raggedright\let\newline\\\arraybackslash\hspace{0pt}}m{#1}}
\newcolumntype{C}[1]{>{\centering\let\newline\\\arraybackslash\hspace{0pt}}m{#1}}
\newcolumntype{R}[1]{>{\raggedleft\let\newline\\\arraybackslash\hspace{0pt}}m{#1}}

\usepackage{microtype}

\aclfinalcopy 

\title{NLP-CUET@LT-EDI-EACL2021: Multilingual Code-Mixed Hope Speech Detection using Cross-lingual Representation Learner}
\author{Eftekhar Hossain{*}, Omar Sharif{\textdagger}  and Mohammed Moshiul Hoque{\textdagger}\\
    {\textdagger}Department of Computer Science and Engineering \\
    {*}Department of Electronics and Telecommunication Engineering\\
    Chittagong University of Engineering and Technology, Bangladesh \\
  \texttt{\{eftekhar.hossain, omar.sharif, moshiul\_240\}@cuet.ac.bd}\\

  }

\date{}

\begin{document}
\maketitle
\begin{abstract}
In recent years, several systems have been developed to regulate the spread of negativity and eliminate aggressive, offensive or abusive contents from the online platforms. Nevertheless, a limited number of researches carried out to identify positive, encouraging and supportive contents. In this work, our goal is to identify whether a social media post/comment contains hope speech or not. We propose three distinct models to identify hope speech in English, Tamil and Malayalam language to serve this purpose. To attain this goal, we employed various machine learning (support vector machine, logistic regression, ensemble), deep learning (convolutional neural network + long short term memory) and transformer (m-BERT, Indic-BERT, XLNet, XLM-Roberta) based methods. Results indicate that XLM-Roberta outdoes all other techniques by gaining a weighted $f_1$-score of $0.93$, $0.60$ and $0.85$ respectively for English, Tamil and Malayalam language. Our team has achieved $1^{st}$, $2^{nd}$ and $1^{st}$ rank in these three tasks respectively.  
\end{abstract}

\section{Introduction}
Nowadays, online and social media platforms have enormous influence and impact on people’s societal life. When people undergo a challenging or unfavourable time, they start to find emotional support from their friends, relatives or even virtual platforms to overwhelm this situation. Due to Covid-19 pandemic, various online forums have become a popular medium of seeking help, suggestion or support. Thus, researchers are trying to develop a computational model that can find positive and supportive social media information. In general, the hope speech contains words of inspiration, promise and suggestions. \citet{chakravarthi-2020-hopeedi} considered those words as hope speech that offer suggestions, reassurance, support, insight and inspiration. Hope speech can be beneficial to save individuals who wish to harm themselves or even attempt to suicide. Such speech inspires people during the period of depression, loneliness and stress with the words of promise, suggestions and support \citep{herrestad2010relational}. The major concern of this research is to originate a computational model on top of this dataset to identify hope speech from the social media posts/comments. Lack of resources on hope speech research, scarcity of training corpora and multilingual code-mixing are the key concerns to develop such models.

Machine learning (ML), and deep learning (DL) based techniques can be utilized to address the problem of hope speech detection. In recent years, transformers have gained immense popularity due to its ability to handle the dependencies between input and output with both attention and recurrence. Consequently, many NLP tasks have accomplished using the transformer-based model to obtain the state-of-the-art performance \citep{chen2021transformer}. The principal contributions in this research as listed below: 
\begin{itemize}
    \item Develop a  model with cross-lingual contextual word embeddings (i.e. transformers) to identify the hope speech considering the code-mixed data for English, Tamil and Malayalam languages.
    \item Investigated the superiority of various ML, DL and transformer-based techniques with detail experimentation.
\end{itemize}

The rest of the paper organized as follows: works related to hope speech detection discussed in Section \ref{section2}. Task and dataset are described in detail in Section \ref{section3}. Section \ref{section4} explains the various techniques used to develop the model for performing the assigned task. Experimental findings and error analysis of the models are introduced in Section \ref{section5}.

\section{Related Work}
\label{section2}
With the substantial growth of the Internet and online contents, several methods have been developed to identify, classify and stop the expansion of negativity such as hate speech detection \citep{mandl2019overview}, hostility detection \citep{sharif2021combating}, aggressive language identification \citep{trac-2020-trolling}, and flagging abusive contents \cite{alw-2020-online}. However, very few researchers have put their focus on the other side that is hope speech detection.  A little work has conducted till to date in this research avenue of NLP. \citet{palakodety12hope} analyzed how hope speech can be utilized to mitigate tension between two rival (Pakistan and India) countries. 
Supporting texts regarding Rohingya community culled from social media in Hindi and English languages \citep{palakodety2020voice}. However, details of the dataset such as inter-annotator agreement, diversity of annotators were not clearly described. \citet{chakravarthi-2020-hopeedi} developed a multilingual code-mixed hope speech dataset for Equality, Diversity and Inclusion (HopeEDI) in English, Tamil and Malayalam language. Data collected from social media like Facebook, YouTube in trending topics, i.e. COVID-19, LGBTIQ issues, and India-China war. Their models achieved the highest weighted $f_1$ score of 0.90, 0.56, 0.70 with a decision tree, naive Bayes, logistic regression techniques for English, Tamil and Malayalam languages.

\section{Task and Dataset Descriptions}
\label{section3}

In this shared task, we have to perform multi-class classification where we aim to identify whether a given comment contains hope speech or not. Our system goal is to classify a post/comment into one of the three predefined classes: hope speech (HS), not hope speech (NHS) and not intended language (NIL). The shared task organizers \citep{dravidianhopespeech-eacl} developed a hope speech corpus in multilingual code-mixed setup. A total of 28541 (for English), 20198 (for Tamil) and 10705 (for Malayalam) texts are available in the corpus. This corpus partitioned into three independent sets: train, validation and test. Initially, the model is developed on top of the train set, and model hyperparameters are tuned based on the validation set's performance. Finally, the model evaluated on the unseen instances of the test set. Table \ref{table1} shows detail statistics of train, validation and test set for each class. 

Further investigation on the training set revealed that the training set is highly imbalanced where several documents in `not intended language' class are much lower than `hope speech' and `not hope speech' classes. The average number of words in `not intended language' is approximately four words for Tamil and Malayalam languages. The model generalization capability on unseen data might degrade due to the lower number of examples on these classes. Detail analysis of the training set presented in table \ref{table2}.

\begin{table*}[h!]
\centering
\begin{tabular}{c|ccc|ccc|ccc}
\hline
 & \multicolumn{3}{c}{\textbf{English}}& \multicolumn{3}{c}{\textbf{Tamil}}& \multicolumn{3}{c}{\textbf{Malayalam}}\\
\hline
&\textbf{HS}&\textbf{NHS}&\textbf{NIL}&\textbf{HS}&\textbf{NHS}&\textbf{NIL}&\textbf{HS}&\textbf{NHS}&\textbf{NIL}\\
\hline
Train & 1962 & 20778 & 22 & 6327 & 7872 &  1961 &  1668  & 6205  & 691\\
Valid &272  & 2569 &  2 & 757 & 998  & 263  & 190 &   784 & 96  \\          
Test & 250  & 2593 &   3  & 815  & 946  & 259 & 194 & 776  & 101  \\               
\hline            
\end{tabular}
\caption{\label{table1} Number of instances in train, validation and test sets for each language. Here HS, NHS and NIL indicates hope speech, not hope speech and not intended language respectively.}
\end{table*}

\begin{table*}[h!]                                                    
\centering
\begin{tabular}{lc|cC{2cm}C{2cm}C{2cm}}
\hline
\textbf{Language}&\textbf{Classes} &\textbf{Total words}&\textbf{Unique words}&\textbf{Max. length (words)}&\textbf{Avg. words (per text)}\\
\hline
\multirow{3}{*}{English} & HS & 49210 & 4811 & 197 & 25.08 \\
&NHS & 317854 & 19740 & 191 & 15.29\\
&NIL& 325 & 239 & 47 &  14.77\\
\cline{2-6}          
\multirow{3}{*}{Tamil}& HS & 56000 & 17274 & 193 & 8.85 \\
&NHS & 76302 & 23977 & 176 & 9.69\\
&NIL& 7309 & 2093 & 48 & 3.72\\
\cline{2-6}
\multirow{3}{*}{Malayalam}& HS & 25144 & 11827 & 96 & 15.07 \\
&NHS & 60313 & 24607 & 95 & 9.72\\
&NIL& 2644 & 1040 & 35 & 3.82\\
\hline
\end{tabular}

\caption{Training set statistics for each language. Here HS, NHS and NIL indicates hope speech, not hope speech and not intended language respectively.}
\label{table2}
\end{table*}


\section{Methodology}
\label{section4}
\begin{figure}[ht!]
\centering
  \includegraphics[width=\linewidth]{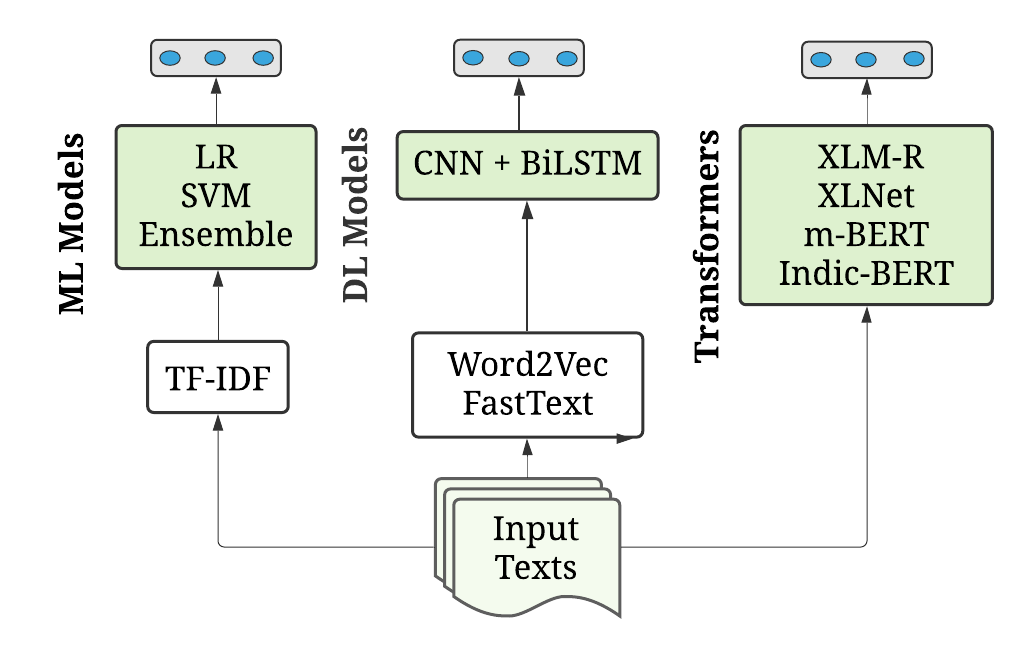}
  \caption{Abstract process of hope speech detection}
  \label{fig:proposed_model}
\end{figure}

This section provides a brief discussion of the schemes and techniques employed to address the task (Section \ref{section3}). Initially, different feature extraction techniques are exploited with machine learning, and deep learning approaches for a baseline evaluation. Moreover, we also applied transformers to obtain a better outcome. Figure \ref{fig:proposed_model} shows the abstract process of hope speech detection. Architectures and parameters of different approaches are described in the following subsections.   

\subsection{Feature Extraction Techniques}
ML and DL based techniques are incapable of processing strings or plain text from the raw forms. Thus, extracting of appropriate or relevant features is a prerequisite to train ML and DL based systems. We utilize TF-IDF \citep{tokunaga1994text}, and FastText \citep{bojanowski2017enriching} techniques for extracting features from the texts.

\paragraph{TF-IDF:}
TF-IDF is a measure that calculates the relevancy of a word to a document in a collection of documents. We calculate the TF-IDF  value of unigram features for all the languages. During the calculation minimum and maximum document frequency value set at 1.

\paragraph{Word Embedding:} Embedding features can capture the semantic meaning of a word. To get embeddings features for all the languages, we used Keras embedding layer with embedding size 100. During the training phase, Keras tries to find the optimal values of the embedding layer's weight matrix by doing simple matrix multiplication and thus create a mapping of each unique words into a vector of real numbers. We utilize the full vocabulary of the corpus and choose maximum input text length $100$, $50$, and $80$ respectively for English, Tamil, and Malayalam data.   

\paragraph{FastText:} To alleviate the problem of out of vocabulary words in keras embeddings, we use FastText embedding. Instead of learning vectors directly for words, FastText represents each word as n-gram of characters. Therefore even if a word was not encountered during training, it could be split into n-gram to get its embedding. Pre-trained \citep{grave2018learning} embedding vectors are used to accomplish the tasks of each language. We retain the default embedding dimension $300$ for FastText embedding.  

\subsection{ML Baselines}
To address the problem, we investigate the performance of three traditional ML approaches, including logistic regression (LR), support vector machine (SVM) and Ensemble. Scikit-learn library is employed for the implementation of these models. TF-IDF features are used to train all the ML methods for three languages. 

\paragraph{LR:} 
LR is constructed by using `lbfgs' solver along with `l2' penalty. The regularization parameter $C$ settled to $2$, $5$, and $1$ respectively for English, Tamil, and Malayalam data.    

\paragraph{SVM:} For SVM, `linear' kernel is utilized with $C$ value of $10$, $1$ and $0.5$ respectively for English, Tamil, and Malayalam language.

\paragraph{Ensemble:}  To perform classification task ensemble approach has proven superior compared to individual models outcome \citep{roy2018ensemble}. We employ decision tree (DT) and random forest (RF) classifiers along with SVM and LR to develop an ensemble method. For RF, $100$ `n\_estimators' is chosen while `gini' criterion used for both DT and RF. On the other hand, previously mentioned parameters have retained in LR and SVM. The majority voting technique is utilized to get the prediction from the ensemble approach.

\begin{table*}[t!]
\centering
\begin{tabular}{cl|ccc|ccc|ccc}
\hline
\textbf{Method}&\textbf{Classifiers} & \multicolumn{3}{c}{\textbf{English}}& \multicolumn{3}{c}{\textbf{Tamil}}& \multicolumn{3}{c}{\textbf{Malayalam}}\\
\hline
&&\textbf{P}&\textbf{R}&\textbf{F}&\textbf{P}&\textbf{R}&\textbf{F}&\textbf{P}&\textbf{R}&\textbf{F}\\
\hline                   
\multirow{3}{*}{ML models}& LR & 0.914 & 0.869 & 0.886  &  0.569 & 0.563 & 0.562 & 0.815 & 0.798 & 0.804\\
&SVM &0.915  &  0.877  & 0.892  & 0.582 & 0.574 & 0.564  & 0.820 & 0.809 & 0.813   \\          
&Ensemble & 0.904 & 0.920 &  0.905  & 0.584 & 0.584 & 0.573 &  0.80 & 0.811 & 0.794 \\                    
\hline            
\multirow{2}{*}{DL models}& C + L (KE) &  0.906 & 0.892 & 0.899 & 0.569 & 0.559  & 0.540 & 0.806 &  0.796 & 0.791  \\              
&C + L (FT) & 0.898 & 0.899 & 0.898 &  0.565 & 0.557 & 0.548  & 0.789 & 0.788 & 0.786   \\       
\hline

\multirow{4}{*}{Transformers}&m-BERT & 0.928 & 0.927 & 0.928 &  0.588 & 0.591 & 0.588 & 0.808 & 0.823 &  0.804  \\                     
&Indic-BERT & 0.913 &  0.920 &   0.910 &  0.593 & 0.592 & 0.578  &  0.839 & 0.842 & 0.840  \\ 
&Xlnet & 0.931 & 0.929 & 0.930  &  0.558 & 0.560 & 0.558 & 0.779 & 0.797 & 0.781 \\      
&XLM-R &  \textbf{0.931} & \textbf{0.931} & \textbf{0.931}  & \textbf{0.610} & \textbf{0.609} & \textbf{0.602}  & \textbf{0.859} & \textbf{0.852} & \textbf{0.854} \\       
\hline
\end{tabular}
\caption{\label{result} Performance comparison of different models on test set where P, R, F denotes precision, recall and weighted $f_1$-score. Here, C+L means the combination of CNN and BiLSTM method and KE and FT represents Keras and FastText embeddings. 
}
\end{table*}

\subsection{DL Baselines} 
A deep learning-based approach is applied with word embedding features to address the task. The model is developed in TensorFlow backend by using Keras library. The combination of convolutional neural network (CNN) and bidirectional long short term memory (BiLSTM) has achieved an outstanding result in many NLP tasks \citep{sharif2020techtexc}. In this approach, we employ one BiLSTM layer on top of a convolution layer. Initially embedding features are feed to the CNN layer consisting of $128$ filters. Following this, to choose appropriate features, a max-pooling is applied with window size $5$. The resultant vector is then passed into the BiLSTM layer. In order to capture long term dependencies, $100$ bidirectional cells are used in this layer. To mitigate the chance of overfitting BiLSTM layer dropout technique is utilized with a dropout rate of $0.2$. Afterwards, the concatenated output of the BiLSTM layer transferred into a softmax layer for the prediction.  

\subsection{Transformers} 
We employed four pre-trained transformer models such as multilingual bidirectional encoder representations from transformers (m-BERT), Indic-BERT, XLNet,  and XLM-Roberta (XLM-R) and fine-tuned them on the dataset with varying hyperparameters. For fine-tuning, maximum text length settled to $50$ for Tamil and $100$ for Malayalam and English. The models are fetched from Huggingface\footnote{ https://huggingface.co/transformers/} transformers library and implemented using Ktrain \citep{maiya2020ktrain} package.      

\paragraph{m-BERT:} m-BERT \citep{devlin2018bert} is pre-trained on a large corpus of multilingual data. To accomplish our purpose, we employed `bert-base-multilingual-cased' model and fine-tuned it on our dataset with batch size $12$.    

\paragraph{Indic-BERT:} Indic-BERT \citep{kakwani2020inlpsuite} is a multilingual model pre-trained specifically on $12$ major Indian languages. It has fewer parameters than other multi-lingual models (i.e. m-BERT, XLM-R). Nevertheless, it outperforms other transformers on various task \citep{kulkarni2021experimental}. The model is fine-tuned with the batch size of $8$. 

\paragraph{XLNet:} XLNet \citep{yang2019xlnet} is an auto-regressive language model which utilizes the recurrence to output the joint probability of a sequence of words. It combines transformer mechanism with slight modification in language modelling approach. For the implementation `xlnet-base-cased' model is used and for all the languages we choose batch size $12$ for fine-tuning.  

\paragraph{XLM-Roberta:} XLM-R \citep{conneau2019unsupervised} is referred as cross lingual representation learner. It is a multi-lingual transformer-based model pre-trained with more that 100 languages and achieves the state-of-the-art performance on cross-lingual NLP tasks. We used `xlm-Roberta-base' model and select a  batch size of $4$ to fine-tuned it on our datasets.    

All transformer models are fine-tuned using Ktrain `fit\_onecycle' method, trained for $30$ epochs with a learning rate of $2e^{-5}$. The early stopping technique is employed to avoid the overfitting problem.  

\section{Results and Analysis}
\label{section5}
This section presents a comprehensive performance analysis of various machine learning, deep learning and transformer models for three languages (English, Tamil and Malayalam). Weighted $f_1$ score uses to determine the excellence of the models. In some cases, other evaluation metrics like precision and recall also considered. Table~\ref{result} presents the evaluation results of all models on the test set. It observed that ensemble achieved the highest $f_1$-score of $0.905$ and $0.573$ respectively for English and Tamil data in ML models. On the other hand, maximum $f_1$-score of $0.813$ is obtained by SVM for the Malayalam language. For all the languages, LR also performed quite similar to SVM but failed to beat other models.

In deep learning, the combination of CNN and BiLSTM experiments with two different embedding features (i.e. Keras embedding and FastText). Both models obtained the highest $f_1$-score of around $0.90$ for English while achieved approximately $0.79$ for Malayalam dataset. However, the combination of CNN-BiLSTM model with FastText embedding features shows $1\%$ rise in $f_1$-score (from $0.54$ to $0.55$) for Tamil language.

On the other hand, Transformer based models showed remarkable performance for all three languages. In English data, m-BERT, XLNet, and XLM-R got the highest $f_1$-score of approximately $0.93$. However, considering both the precision and recall values, only XLM-R outperformed the other models. For the Tamil language, $f_1$-score of around $0.56$ and $0.58$ respectively obtained by XLNet and Indic-BERT. A slight rise of $1\%$ in $f_1$-score ($0.588$) is noticed for m-BERT but it cannot beat XLM-R performance ($f_1$ score = $0.602$). In case of Malayalam data, Indic-BERT ($f_1$ score = $0.84$) shows an increase of  $4\%$ to $6\%$ than the $f_1$-score obtained by m-BERT ($0.804$) and XLNet ($0.781$). Nevertheless, it failed to reach the outcome of XLM-R ($f_1$ score = $0.854$), which outdoes all the models.

The results show that \textbf{XLM-R} outperformed ML, DL, and other transformer-based models for all three languages. The ability of cross-lingual understanding at different linguistic levels might be the reason for this superior performance of XLM-R.  

\begin{figure*}[t!]
\begin{multicols}{3}
    \subfigure[]{\includegraphics[height=4cm, width=0.33\textwidth]{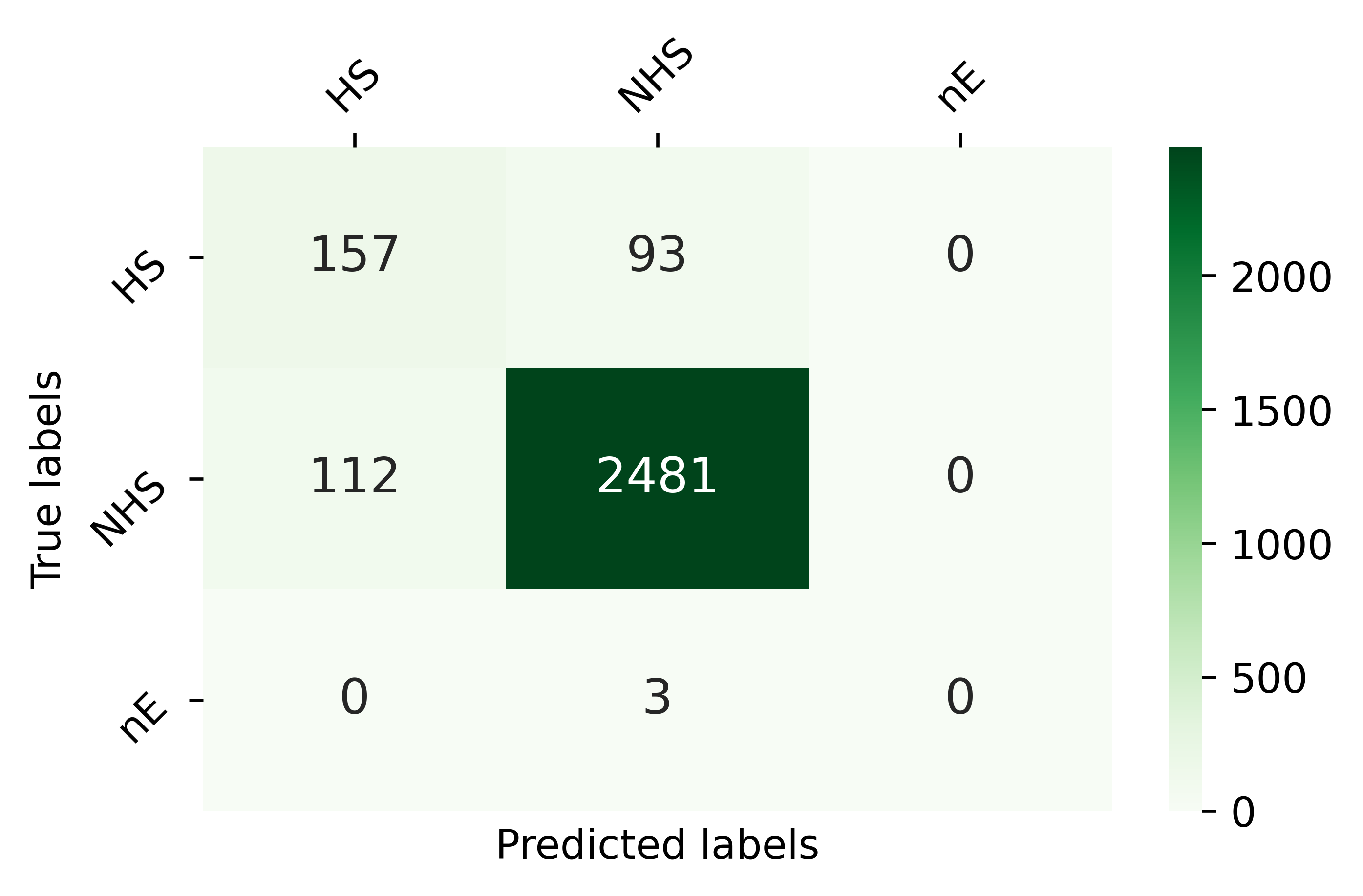}}
    \subfigure[]{\includegraphics[height=4cm, width=0.33\textwidth]{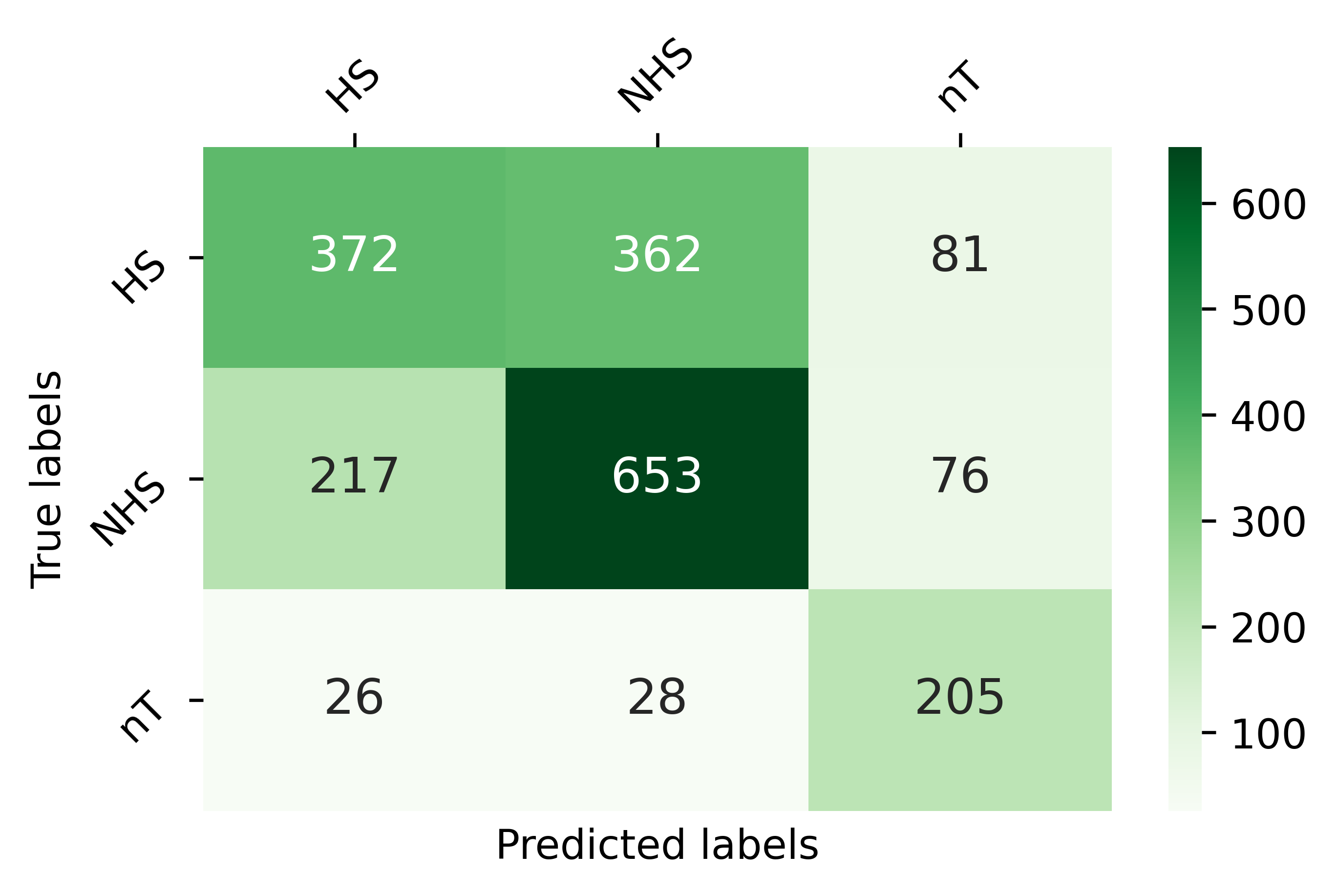}}
    \subfigure[]{\includegraphics[height=4cm, width=0.33\textwidth]{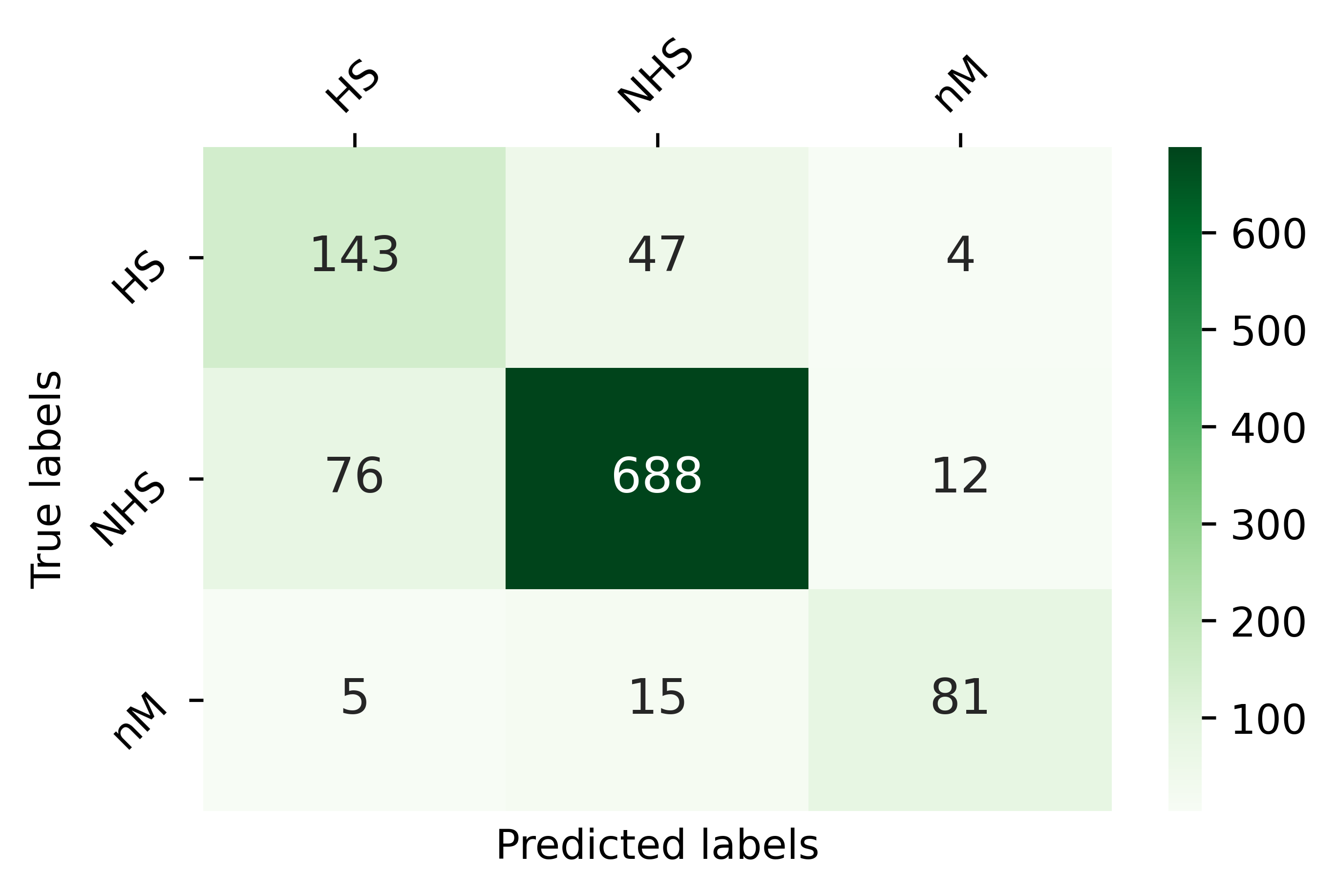}}
   
    \end{multicols}
 \caption{Confusion matrix of XLM-R technique for (a) English, (b) Tamil and (c) Malayalam languages.}
 \label{confusion}
\end{figure*}

\subsection{Error Analysis}
It is evident from the Table~\ref{result} that XLM-R is the best performing model to detect hope speech for English, Tamil and Malayalam languages. A detail error analysis is carried out using the confusion matrix to investigate more insights concerning the individual class performance (Figure~\ref{confusion}).
From the Figure~\ref{confusion}(a), it is noticed that among $250$ HS instances, $93$ are misclassified as NHS. However, the model ultimately failed to detect any of the not-English data and incorrectly classified them as NHS. Similarly, $81$ (out of $815$) HS and $76$ (out of 946) NHS are misclassified as a not-Tamil class in the Tamil language. However, the misclassification rate is comparatively low for nT class. Figure~\ref{confusion}(c) shows that the model correctly classified $143$ HS (out of $194$) instances. Moreover, it wrongly classified $47$ samples (as NHS) and $4$ samples (as nM). Likewise, among $101$ instances of NM, the model misclassified $20$ texts while correctly identifying $ 81$ instances. On the other hand, the NHS class received $668$ correct classification out of $756$ instances, and only misclassified $88$ instances as other classes.

We noticed that the model mostly gets confused with HS and NHS class in all languages from the error analysis. The possible reason is that there may be plenty of code-mixed words common in both classes. Thus the system could not apprehend the inherent meaning of the sentences. The high-class imbalance may be another likely reason why the model gives the most priority to the not hope speech class and therefore incorrectly classified hope speech as not hope speech. Increasing the number of instances in the NHS class may mitigate the chance of excessive misclassification.     

\section{Conclusion}
This paper describes and analyses the several ML, DL, and transformer-based methods that we have adopted to participate in the hope speech detection shared task at EACL 2021. Employing TF-IDF, embedding features initially, we performed experiments with ML (LR, SVM, ensemble) and DL (CNN+BiLSTM) approaches. The outcome shows that the ensemble technique achieved higher performance compared to other ML/DL models. Further, transformer-based techniques are employed to improve the overall performance. The XML-R model outperformed all the models' performance by achieving the highest weighted $f_1$-score of $0.931$, $0.854$, and $0.602$ respectively for English, Tamil Malayalam language. In the future, contextualized embeddings (such as ELMO, FLAIR) and transformers ensemble might explore to investigate the system's performance.

\bibliography{anthology,eacl2021}

\begin{thebibliography}{21}
\expandafter\ifx\csname natexlab\endcsname\relax\def\natexlab#1{#1}\fi

\bibitem[{Akiwowo et~al.(2020)Akiwowo, Vidgen, Prabhakaran, and
  Waseem}]{alw-2020-online}
Seyi Akiwowo, Bertie Vidgen, Vinodkumar Prabhakaran, and Zeerak Waseem,
  editors. 2020.
\newblock \href {https://www.aclweb.org/anthology/2020.alw-1.0}
  {\emph{Proceedings of the Fourth Workshop on Online Abuse and Harms}}.
  Association for Computational Linguistics, Online.

\bibitem[{Bojanowski et~al.(2016)Bojanowski, Grave, Joulin, and
  Mikolov}]{bojanowski2017enriching}
Piotr Bojanowski, Edouard Grave, Armand Joulin, and Tom{\'{a}}s Mikolov. 2016.
\newblock \href {http://arxiv.org/abs/1607.04606} {Enriching word vectors with
  subword information}.
\newblock \emph{CoRR}, abs/1607.04606.

\bibitem[{Chakravarthi(2020)}]{chakravarthi-2020-hopeedi}
Bharathi~Raja Chakravarthi. 2020.
\newblock \href {https://www.aclweb.org/anthology/2020.peoples-1.5}
  {{H}ope{EDI}: A multilingual hope speech detection dataset for equality,
  diversity, and inclusion}.
\newblock In \emph{Proceedings of the Third Workshop on Computational Modeling
  of People's Opinions, Personality, and Emotion's in Social Media}, pages
  41--53, Barcelona, Spain (Online). Association for Computational Linguistics.

\bibitem[{Chakravarthi and Muralidaran(2021)}]{dravidianhopespeech-eacl}
Bharathi~Raja Chakravarthi and Vigneshwaran Muralidaran. 2021.
\newblock Findings of the shared task on {H}ope {S}peech {D}etection for
  {E}quality, {D}iversity, and {I}nclusion.
\newblock In \emph{Proceedings of the First Workshop on Language Technology for
  Equality, Diversity and Inclusion}. Association for Computational
  Linguistics.

\bibitem[{Chen et~al.(2021)Chen, Chen, Gao, Chen, Huo, Meng, Ren, and
  Zhou}]{chen2021transformer}
Ben Chen, Bin Chen, Dehong Gao, Qijin Chen, Chengfu Huo, Xiaonan Meng, Weijun
  Ren, and Yang Zhou. 2021.
\newblock \href {http://arxiv.org/abs/2101.05509} {Transformer-based language
  model fine-tuning methods for covid-19 fake news detection}.

\bibitem[{Conneau et~al.(2019)Conneau, Khandelwal, Goyal, Chaudhary, Wenzek,
  Guzm{\'{a}}n, Grave, Ott, Zettlemoyer, and
  Stoyanov}]{conneau2019unsupervised}
Alexis Conneau, Kartikay Khandelwal, Naman Goyal, Vishrav Chaudhary, Guillaume
  Wenzek, Francisco Guzm{\'{a}}n, Edouard Grave, Myle Ott, Luke Zettlemoyer,
  and Veselin Stoyanov. 2019.
\newblock \href {http://arxiv.org/abs/1911.02116} {Unsupervised cross-lingual
  representation learning at scale}.
\newblock \emph{CoRR}, abs/1911.02116.

\bibitem[{Devlin et~al.(2018)Devlin, Chang, Lee, and
  Toutanova}]{devlin2018bert}
Jacob Devlin, Ming{-}Wei Chang, Kenton Lee, and Kristina Toutanova. 2018.
\newblock \href {http://arxiv.org/abs/1810.04805} {{BERT:} pre-training of deep
  bidirectional transformers for language understanding}.
\newblock \emph{CoRR}, abs/1810.04805.

\bibitem[{Grave et~al.(2018)Grave, Bojanowski, Gupta, Joulin, and
  Mikolov}]{grave2018learning}
Edouard Grave, Piotr Bojanowski, Prakhar Gupta, Armand Joulin, and Tom{\'{a}}s
  Mikolov. 2018.
\newblock \href {http://arxiv.org/abs/1802.06893} {Learning word vectors for
  157 languages}.
\newblock \emph{CoRR}, abs/1802.06893.

\bibitem[{Herrestad and Biong(2010)}]{herrestad2010relational}
Henning Herrestad and Stian Biong. 2010.
\newblock Relational hopes: A study of the lived experience of hope in some
  patients hospitalized for intentional self-harm.
\newblock \emph{International journal of qualitative studies on health and
  well-being}, 5(1):4651.

\bibitem[{Kakwani et~al.(2020)Kakwani, Kunchukuttan, Golla, N.C.,
  Bhattacharyya, Khapra, and Kumar}]{kakwani2020inlpsuite}
Divyanshu Kakwani, Anoop Kunchukuttan, Satish Golla, Gokul N.C., Avik
  Bhattacharyya, Mitesh~M. Khapra, and Pratyush Kumar. 2020.
\newblock \href {https://doi.org/10.18653/v1/2020.findings-emnlp.445}
  {{I}ndic{NLPS}uite: Monolingual corpora, evaluation benchmarks and
  pre-trained multilingual language models for {I}ndian languages}.
\newblock In \emph{Findings of the Association for Computational Linguistics:
  EMNLP 2020}, pages 4948--4961, Online. Association for Computational
  Linguistics.

\bibitem[{Kulkarni et~al.(2021)Kulkarni, Mandhane, Likhitkar, Kshirsagar,
  Jagdale, and Joshi}]{kulkarni2021experimental}
Atharva Kulkarni, Meet Mandhane, Manali Likhitkar, Gayatri Kshirsagar,
  Jayashree Jagdale, and Raviraj Joshi. 2021.
\newblock \href {http://arxiv.org/abs/2101.04899} {Experimental evaluation of
  deep learning models for marathi text classification}.

\bibitem[{Kumar et~al.(2020)Kumar, Ojha, Lahiri, Zampieri, Malmasi, Murdock,
  and Kadar}]{trac-2020-trolling}
Ritesh Kumar, Atul~Kr. Ojha, Bornini Lahiri, Marcos Zampieri, Shervin Malmasi,
  Vanessa Murdock, and Daniel Kadar, editors. 2020.
\newblock \href {https://www.aclweb.org/anthology/2020.trac-1.0}
  {\emph{Proceedings of the Second Workshop on Trolling, Aggression and
  Cyberbullying}}. European Language Resources Association (ELRA), Marseille,
  France.

\bibitem[{Maiya(2020)}]{maiya2020ktrain}
Arun~S. Maiya. 2020.
\newblock \href {http://arxiv.org/abs/2004.10703} {ktrain: A low-code library
  for augmented machine learning}.

\bibitem[{Mandl et~al.(2019)Mandl, Modha, Majumder, Patel, Dave, Mandlia, and
  Patel}]{mandl2019overview}
Thomas Mandl, Sandip Modha, Prasenjit Majumder, Daksh Patel, Mohana Dave,
  Chintak Mandlia, and Aditya Patel. 2019.
\newblock Overview of the hasoc track at fire 2019: Hate speech and offensive
  content identification in indo-european languages.
\newblock In \emph{Proceedings of the 11th Forum for Information Retrieval
  Evaluation}, pages 14--17.

\bibitem[{Palakodety et~al.(2019{\natexlab{a}})Palakodety, KhudaBukhsh, and
  Carbonell}]{palakodety12hope}
Shriphani Palakodety, Ashiqur~R. KhudaBukhsh, and Jaime~G. Carbonell.
  2019{\natexlab{a}}.
\newblock \href {http://arxiv.org/abs/1909.12940} {Kashmir: {A} computational
  analysis of the voice of peace}.
\newblock \emph{CoRR}, abs/1909.12940.

\bibitem[{Palakodety et~al.(2019{\natexlab{b}})Palakodety, KhudaBukhsh, and
  Carbonell}]{palakodety2020voice}
Shriphani Palakodety, Ashiqur~R. KhudaBukhsh, and Jaime~G. Carbonell.
  2019{\natexlab{b}}.
\newblock \href {http://arxiv.org/abs/1910.03206} {Voice for the voiceless:
  Active sampling to detect comments supporting the rohingyas}.
\newblock \emph{CoRR}, abs/1910.03206.

\bibitem[{Roy et~al.(2018)Roy, Kapil, Basak, and Ekbal}]{roy2018ensemble}
Arjun Roy, Prashant Kapil, Kingshuk Basak, and Asif Ekbal. 2018.
\newblock \href {https://www.aclweb.org/anthology/W18-4408} {An ensemble
  approach for aggression identification in {E}nglish and {H}indi text}.
\newblock In \emph{Proceedings of the First Workshop on Trolling, Aggression
  and Cyberbullying ({TRAC}-2018)}, pages 66--73, Santa Fe, New Mexico, USA.
  Association for Computational Linguistics.

\bibitem[{Sharif et~al.(2020)Sharif, Hossain, and Hoque}]{sharif2020techtexc}
Omar Sharif, Eftekhar Hossain, and Mohammed~Moshiul Hoque. 2020.
\newblock \href {http://arxiv.org/abs/2012.11420} {Techtexc: Classification of
  technical texts using convolution and bidirectional long short term memory
  network}.

\bibitem[{Sharif et~al.(2021)Sharif, Hossain, and Hoque}]{sharif2021combating}
Omar Sharif, Eftekhar Hossain, and Mohammed~Moshiul Hoque. 2021.
\newblock \href {http://arxiv.org/abs/2101.03291} {Combating hostility:
  Covid-19 fake news and hostile post detection in social media}.

\bibitem[{Tokunaga and Makoto(1994)}]{tokunaga1994text}
Takenobu Tokunaga and Iwayama Makoto. 1994.
\newblock Text categorization based on weighted inverse document frequency.
\newblock In \emph{Special Interest Groups and Information Process Society of
  Japan (SIG-IPSJ}. Citeseer.

\bibitem[{Yang et~al.(2019)Yang, Dai, Yang, Carbonell, Salakhutdinov, and
  Le}]{yang2019xlnet}
Zhilin Yang, Zihang Dai, Yiming Yang, Jaime~G. Carbonell, Ruslan Salakhutdinov,
  and Quoc~V. Le. 2019.
\newblock \href {http://arxiv.org/abs/1906.08237} {Xlnet: Generalized
  autoregressive pretraining for language understanding}.
\newblock \emph{CoRR}, abs/1906.08237.

\end{thebibliography}
\bibliographystyle{acl_natbib}

\end{document}